\documentclass[letterpaper,english]{ieeeconf}
\usepackage[T1]{fontenc}
\usepackage[latin9]{inputenc}
\usepackage{color}
\usepackage{refstyle}
\usepackage{amsmath}
\usepackage{amssymb}
\usepackage{graphicx}
\usepackage{wrapfig}

\makeatletter

\AtBeginDocument{\providecommand\Figref[1]{\ref{fig:#1}}}
\AtBeginDocument{\providecommand\Eqref[1]{\ref{Eq:#1}}}
\AtBeginDocument{}
\AtBeginDocument{\providecommand\Figref[1]{\ref{Fig:#1}}}
\AtBeginDocument{\providecommand\eqref[1]{\ref{eq:#1}}}
\AtBeginDocument{}
\pdfpageheight\paperheight
\pdfpagewidth\paperwidth

\RS@ifundefined{subsecref}
  {\newref{subsec}{name = \RSsectxt}}
  {}
\RS@ifundefined{thmref}
  {\def\RSthmtxt{theorem~}\newref{thm}{name = \RSthmtxt}}
  {}
\RS@ifundefined{lemref}
  {\def\RSlemtxt{lemma~}\newref{lem}{name = \RSlemtxt}}
  {}

\usepackage{algpseudocode}
\usepackage{xfrac}

\usepackage{cite}

\makeatother

\usepackage{babel}
\begin{document}
\title{{SplitFlyer: a Modular Quadcoptor that Disassembles into Two Flying Robots}}
\author{Songnan Bai$^\dagger$, Shixin Tan$^\dagger$, and Pakpong Chirarattananon 
\thanks{This work was supported by the Research Grants Council of the Hong Kong Special Administrative Region of China (grant number CityU-11207718)}
\thanks{$^\dagger$These authors contributed equally to this work.}
\thanks{The authors are with the Department of Biomedical Engineering, City University of Hong Kong, Hong Kong SAR, China (email: pakpong.c@cityu.edu.hk).} }
\maketitle
\begin{abstract}
We introduce SplitFlyer--a novel quadcopter with an ability to disassemble into two self-contained bicopters through human assistance. As a subunit, the bicopter is a severely underactuated aerial vehicle equipped with only two propellers. Still, each bicopter is capable of independent flight. To achieve this, we provide an analysis of the system dynamics by relaxing the control over the yaw rotation, allowing the bicopter to maintain its large spinning rate in flight. Taking into account the gyroscopic motion, the dynamics are described and a cascaded control strategy is developed. We constructed a transformable prototype to demonstrate consecutive flights in both configurations. The results verify the proposed control strategy and show the potential of the platform for future research in modular aerial swarm robotics.
\end{abstract}

\section{Introduction}
In recent years, there has been rapid advances in the field of micro aerial vehicles (MAVs). Multicopters, in particular, receive tremendous attentions owing to their simplicity \cite{mellinger2012trajectory}, agility \cite{faessler2017differential}, and versatility \cite{mu2019universal,shu2019quadrotor}. To accommodate a wide range of possible applications, research in MAVs has branched into several related areas, including aerial manipulation \cite{ryll20196d}, collective behavior \cite{mcguire2019minimal}, multimodal locomotion \cite{hsiao2019ceiling}, and modular and reconfigurable robotics \cite{mu2019universal,saldana2018modquad,anzai2019design,win2019dynamics}.

This work is motivated by the potential of swarm behaviors and modular robotics. As demonstrated by biological systems such as ants and bee colonies, swarm intelligence allows individuals with limited capabilities to collectively accomplish complex tasks. Similarly, uses of modularity in robotics expand functionality by letting systems adapt its form and locomotion through reconfiguration or self-(dis)assembly \cite{seo2019modular,yao2019reconfiguration,savoie2019robot}. In aerial robotics, modularity has been employed to either allow a robot to be constructed from a flightless base module \cite{mu2019universal,anzai2019design} or bring together several flight-capable vehicles for mid-air self-assembly \cite{saldana2018modquad,oung2014distributed}.
\begin{figure}
\begin{centering}
\includegraphics{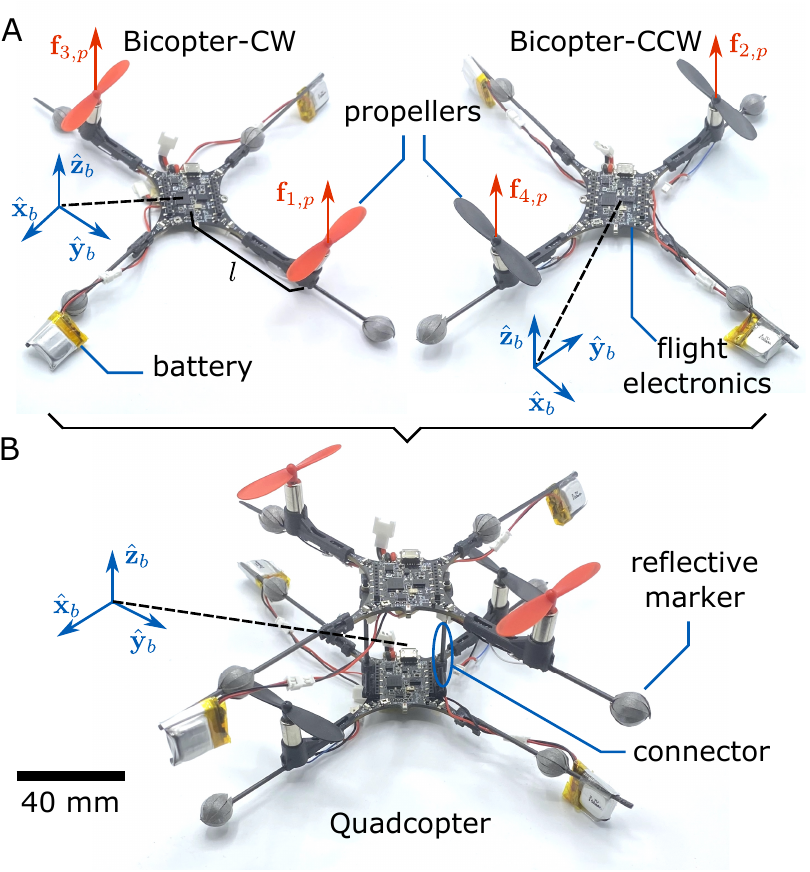}
\par\end{centering}
\caption{Photograhs of the proposed aerial robot. (A) Bicopter mode. SplitFlyer is made up of two flight-capable submodules distinguished by the spinning direction of the propellers, Bicopter-CW and Bicopter-CCW. (B) Quadcopter mode. A complete robot is constructed from two bicopter modules. In this configuration, the robot operates as a conventional multirotor vehicle.\label{fig:design}}
\end{figure}

This paper presents a novel modular aerial vehicle--SplitFlyer. In the original quadcopter form (\Figref{design}B), the robot resembles a regular multirotor platform with four propellers minimally required to attain conventional hovering flights. Nevertheless, through a simple human assistance, the robot is dismantled into two self-contained bicopters, each with the ability to fly independently despite possessing only two actuators. With future development, this conceptual prototype will be equipped to autonomously disassemble mid-air. When deployed in large number, SplitFlyers have an ability to double the flock size, boosting their potential in search and rescue missions or other swarm applications.

Unlike tamden rotors with controllable blade pitches or a bicopter with added servomotors \cite{qin2020gemini}, the disassembled robot in the bicopter form is equipped with only two motors and is severely underactuated. This brings associated challenges in flight control and stability. Few researchers have proposed a strategy to model and control similar underactuated multirotor vehicles \cite{zhang2016controllable,mueller2016relaxed} by introducing the concept of relaxed hovering solutions. In \cite{zhang2016controllable,mueller2016relaxed}, the authors propose a framework to identify a periodic solution of translational and rotational dynamics. By linearizing the system around those relaxed hovering solutions, the controllability is verified. This allows a controller to be derived with linear system methods such as LQR. The strategy can be generally applied to multirotor systems with one, two, or three propellers.

Herein, we take a different approach to model the flight dynamics and secure flight stability.  Leveraging the symmetry of the  bicopters, our method considers the angular momentum of the spinning robot to derive the equations of motion taking into account its gyroscopic motion. Through some simplifying assumptions, a flight controller with cascaded structure is developed. \textcolor{black}{The controller incorporates aerodynamic damping caused by the fast yaw rotation, taking into account both instantaneous and cycle-averaged dynamics to allow the aerial robot with only two actuators to be stable and position controlled.} The physics-based method benefits from the gained insights, demonstrating the role of aerodynamic drag and fast yaw rotation on the attitude stability.

This  paper  is  organized  as  follows.  Section  II,  discusses the system architecture. This is followed by the analysis of flight dynamics and control strategy of a bicopter in Section III. The quadcopter mode is briefly explained in Section IV. Flight experiments of the robot in both configurations, including the transformation, are reported in Section V. Lastly, conclusion and future directions are provided.



\section{Split Quadcopter Design\label{sec:Split-Quadcopter-Design}}

SplitFlyer is an modular robot with two modes of aerial locomotion. The robot is composed of two flight-capable bicopters as base units. Through human assistance, the two modules can be reconfigured into a conventional quadcopter as illustrated in \Figref{design}.

In the bicopter configuration (\Figref{design}A), each robot consists of its own airframe, flight electronics, onboard batteries, and a pair of motors and propellers. Two small batteries are incorporated and strategically placed in each bicopter to achieve the desired mass distribution (see Section \ref{subsec:gyroscopic}). The inclusion of all essential components permits both base units to fly independently and the human-assisted reconfiguration is solely mechanical (no electrical connections required between two units). As a result, two bicopters are nearly identical, except for the propellers' spinning directions. We employ the annotations CW and CCW to distinguish the module according to the direction of the propeller's torque, with CW corresponding to the unit with propelling thrust and torque aligned.

With only two actuators, the bicopter is severely underactuated. The robot is still capable of achieving stable trajectory following flights or hovering by foregoing the independent control of the yaw rate as described as a relaxed hover condition in \cite{mueller2016relaxed,zhang2016controllable}. As a consequence of the non-zero yaw torque generated by the spinning propellers, the robot flies with a relatively high yaw velocity. Therefore, the flight dynamics and control are derived with the consideration of the gyroscopic effect.

In the combined configuration (\Figref{design} (A)), the resultant quadcopter takes a configuration resembling a conventional multirotor. The vertical offset between two propeller pairs does not directly affect its flight dynamics \cite{mu2019universal}, allowing existing analysis and control methods to be used. 

To support both flight modes, the flight controllers are programmed to automatically detect the current flight configuration by exploiting the difference between two flight mechanisms. This is obtained by monitoring the yaw rate during the takeoff period, permitting the controller to rapidly activate the correct control method without requiring an electrical connection between the modules.


\begin{figure}
\centering{}\includegraphics{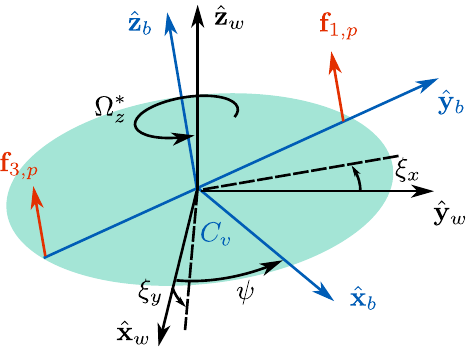}\caption{A schematic diagram illustrating the flight dynamics of a bicopter. The inertial frame ($\hat{\mathbf{x}}_w\hat{\mathbf{y}}_w\hat{\mathbf{z}}_w$) is falsely drawn on top of the body frame ($\hat{\mathbf{x}}_b\hat{\mathbf{y}}_b\hat{\mathbf{z}}_b$) for clarity. The disk represents the plane of the robot's yaw rotation, coinciding with the $\hat{\mathbf{x}}_{b}$-$\hat{\mathbf{y}}_{b}$ plane.
\label{fig:Schematics}}
\end{figure}

\section{Bicopter Flight}

In this work, we regard a bicopter as a modular flight-capable self-contained unit. Two bicopters with opposite spinning directions constitute sufficient components to form a complete conventional quadcopter. With two propellers, the bicopter inevitably creates substantial yaw torque in operation. This calls for a thorough investigation into its distinct flight dynamics and the development of a compatible control framework.

\subsection{Translational dynamics\label{subsec:Dynamics33}}

To describe the bicopter translational dynamics, we let $\mathbf{p}=\left[x,y,z\right]^{\mathrm{T}}$ denote the position of a bicopter in the inertial frame ($\hat{\mathbf{x}}_w\hat{\mathbf{y}}_w\hat{\mathbf{z}}_w$), $m$ be the mass, and $g$ be the gravitational constant. The equation of motion is
\begin{equation}
m\ddot{\mathbf{p}}={}^{b}\mathbf{R}_{w}\mathop{\sum_{i}}\mathbf{f}_{i}-mg\mathbf{e}_3,\label{eq:transldynamic}
\end{equation}
where, the summation is for $i={1,3}$ for Bicopter-CW and $i={2,4}$ for Bicopter-CCW,  $^{b}\mathbf{R}_{w}$ is a rotation matrix mapping the robot's frame to the inertial frame, $\mathbf{e}_i$ is a basis vector, and $\mathbf{f}_{i}$ is the force associated with the $i^{\text{th}}$ propeller, often taken as the aerodynamic thrust generated by the propeller ($\mathbf{f}_{i,p}$) \cite{mu2019universal}. However, in instances where a propeller moves at significant speed with respect to still air, it brings about additional drag such that
\begin{equation}
\mathbf{f}_{i}=\mathbf{f}_{i,p}+\mathbf{f}_{i,d}. \label{eq:prop_force}
\end{equation}
The rotor drag can be approximated a linear function of the local air velocity perceived by each propeller ($\mathbf{v}_i$): $\mathbf{f}_{i,d}=-\mathbf{D}\mathbf{v}_i$, where $\mathbf{D}=\text{diag}(D_h,D_h,D_v)$ is a constant diagonal matrix representing the drag coefficient \cite{faessler2017differential}. $\mathbf{v}_i$ is computed from the combination of $\dot{\mathbf{p}}$ and the vehicle's angular velocity with respect to the body frame ($\mathbf{\Omega}=[\Omega_x,\Omega_y,\Omega_z]^T$) as $\mathbf{v}_{i}={}^{w}\mathbf{R}_{b}\dot{\mathbf{p}}+\mathbf{\mathbf{\Omega}}\times\mathbf{l}_{i}$ where  $\mathbf{l}_{i}$ represents the location of the $i^{\text{th}}$ propeller in the body frame ($l\mathbf{e}_2$ or $-l\mathbf{e}_2$ according to \Figref{design}A). 

Focusing on low-speed, near hovering flights, with the fact that each bicopter has only two propellers, its unique attitude dynamics render the angular velocity term dominates (see Secion \ref{subsec:relaxed_hov_cond} below). As a result, the rotor drag reduces to
\begin{equation}
\mathbf{f}_{i,d}= -\mathbf{D}\left[\begin{array}{ccc}
-\sigma_i \Omega_z l & 0 & \sigma_i \Omega_x l\end{array}\right]^{\mathrm{T}},\label{eq:rotordrag}
\end{equation}
where $\sigma_i=1$ for $i={1,2}$ or $\sigma_i=-1$ for $i={3,4}$.
It can be seen that, in the context of the entire robot, $\sum_i \mathbf{f}_{i,d}=0$. Furthermore, by expressing $\mathbf{f}_{i,p}$ as $f_i\mathbf{e}_3$, \eqref{transldynamic} becomes
\begin{equation}
m\ddot{\mathbf{p}}=\hat{\mathbf{z}}_{b}\mathop{\sum_{i}}f_{i,p}-mg\mathbf{e}_3.\label{eq:transdynamics2}
\end{equation}
\Eqref{transdynamics2} implies that the translational dynamics of the bicopter is chiefly unaffected by the impact of the rotor drag from the high yaw rate. The translational dynamics remain governed by the total thrust and the attitude ($\hat{\mathbf{z}}_b$ of the robot, identical to conventional multirotor vehicles).

\subsection{Attitude dynamics}

As a rigid body, the bicopter's attitude dynamics are provided by the Euler's equations
\begin{equation}
\mathbf{I}\dot{\mathbf{\Omega}}+\mathbf{\Omega}\times\mathbf{I}\mathbf{\Omega}=\sum_i\tau_i, \label{eq:eulers_eq}
\end{equation}
where, $\mathbf{I}$ is the inertia moment, and, similar to  $\mathbf{f}_i$, $\tau_i$ can be written as the sum of thrust-induced torque and drag torque: $\tau_i=\tau_{i,p}+\tau_{i,d}$ in the body frame. More specifically,
\begin{equation}
\tau_{i,p} = \left[\begin{array}{ccc}
\sigma_i l f_{i,p} & 0 & \delta_i c f_{i,p}\end{array}\right]^{\mathrm{T}} ,\label{eq:prop_torque}
\end{equation}
where $\delta_i=(-1)^{i-1}$ distinguishes the difference between Bicopter-CW and Bicopter-CCW and $c$ is a coefficient of the propeller denoting the ratio of torque to thrust. Meanwhile, the drag torque can be computed as
\begin{equation}
\tau_{i,d} =\mathbf{l}_i\times\mathbf{f}_{i,d} = -l^2 \left[\begin{array}{ccc}
D_v \Omega_x & 0 & D_h \Omega_z  \end{array}\right]^{\mathrm{T}} ,\label{eq:drag_torque}
\end{equation}
where we have used the definition of $\mathbf{f}_{i,d}$ from \eqref{rotordrag} and the fact that $\mathbf{l}_i=\sigma_i l\mathbf{e}_2$.

\subsection{Relaxed hovering condition}\label{subsec:relaxed_hov_cond}

To provide further insights, we consider the conditions for an equilibrium flight. According to \eqref{transdynamics2}, the equilibrium condition (denoted with $\cdot^*$) for the translational dynamics is
\begin{equation}
\hat{\mathbf{z}}_b^*=\mathbf{e}_3 \quad \text{and} \quad \mathop{\sum_{i}}f_{i,p}^*=mg. \label{eq:translation_equilibrium}
\end{equation}
The first condition requires $\Omega_x^*=\Omega_y^*=0$. This subsequently restricts the equilibrium state imposed by equations~(\ref{eq:eulers_eq})-(\ref{eq:drag_torque}) to    $\mathop{\sum_{i}}\delta_i c f^*_i = \delta_i c mg = l^2D_h\Omega_z^*$.

In other words, the bicopter nominally stays upright with a constant yaw velocity determined by
\begin{equation}
\Omega^*_z=\delta_i c mg/D_h l^2. \label{eq:attitude_equilibrium}
\end{equation}
The \textit{relaxed} hovering condition ($\Omega_z^*\neq 0$) \cite{mueller2016relaxed} is a consequence of the non-zero yaw torque generated by the bicopter.

\subsection{Gyroscopic motion and reduced attitude dynamics\label{subsec:gyroscopic}}

Despite the non-zero nominal angular velocity, \eqref{transdynamics2} suggests only $\hat{\mathbf{z}}_b$, not the entire attitude, affects the robot's translational motion. In addition, \eqref{attitude_equilibrium} implies a potentially high yaw rate in flight. These motivate us to regard the bicopter as a gyroscope instead of using the full attitude dynamics from \eqref{eulers_eq} for flight stability analysis and control purposes.

Under the assumption of small deviations from the equilibrium state (\eqref{translation_equilibrium}), we let $\hat{\mathbf{z}}_b\approx[\xi_y,-\xi_x,1]^\mathrm{T}$(with  $\left|\xi_{x}\right|,\left|\xi_{y}\right|\ll1$) represent the reduced attitude (see \Figref{Schematics}). Moreover, by design, the vehicle's mass is distributed such that its moment of inertia about the pitch and roll axes are approximately equal or $\mathbf{I}=\mbox{diag}(I_d,I_d,I_z)$. This, with the fact that  $\left|\Omega_{z}^{*}\right|\gg\left|\Omega_{x}\right|,\left|\Omega_{y}\right|$, allows the robot to be treated as an axisymmetric gyroscope with a constant spinning speed $\Omega^*_z$. This means the angular momentum vector of the bicopter defined in the inertial frame is \cite{Crandall1985effect}
\begin{equation}
\mathbf{L}= I_d\dot{\xi}_x \hat{\mathbf{x}}_m +  I_d\dot{\xi}_y \hat{\mathbf{y}}_m  +  I_z\Omega^*_z \hat{\mathbf{z}}_b,\label{eq:angular_momentum}
\end{equation}where $\hat{\mathbf{x}}_m\approx [1,0,-\xi_y]^{\mathrm{T}}$ and $\hat{\mathbf{y}}_m\approx [0,1,\xi_x]^{\mathrm{T}}$ are unit vectors along the axes of the moving frame ($\hat{\mathbf{x}}_m\hat{\mathbf{y}}_m\hat{\mathbf{z}}_b$) as shown in \Figref{Schematics}. Using the fact that $\dot{\hat{\mathbf{x}}}_m=-\dot{\xi}_y\hat{\mathbf{z}}_b$, $\dot{\hat{\mathbf{y}}}_m=\dot{\xi}_x\hat{\mathbf{z}}_b$, and $\dot{\hat{\mathbf{z}}}_b=\dot{\xi}_y\hat{\mathbf{x}}_m-\dot{\xi}_x\hat{\mathbf{y}}_m$, the reduced attitude dynamics are obtained by taking the time derivative of \eqref{angular_momentum} \cite{Crandall1985effect}:
\begin{equation}
I_{d}\ddot{\mathbf{\xi}}+I_{z}\Omega^*_{z}\left[\begin{array}{cc}
0 & 1\\
-1 & 0
\end{array}\right]\dot{\mathbf{\xi}}=\bar{\tau},\label{eq:EoM}
\end{equation}
where $\mathbf{\xi}=[\xi_x,\xi_y]^{\mathrm{T}}$ and $\bar{\tau}$ is taken from the $x$ and $y$ components of the collective torque with respect to the inertial frame ($\bar{\tau}=[\mathbf{e}_1,\mathbf{e}_2]^{\mathrm{T}}\mathbf{R}\sum_i\tau_{i,p}+\tau_{i,d}$).
Near the relaxed hovering state (the bicopter is nearly upright), if we employ roll ($\left|\phi\right|\ll 1$), pitch ($\left|\theta\right|\ll 1$), and yaw angles ($\psi$, defined as the angle between $\hat{\mathbf{x}}_w$ and $\hat{\mathbf{x}}_b$ as depicted in \Figref{Schematics}) to represent $\mathbf{R}$, it can be shown that $\Omega_x\approx\dot{\phi}\approx\cos \psi\dot{\xi}_x+\sin\psi \dot{\xi}_y$. Substituting this into \eqref{drag_torque}, keeping only first-order terms, the outcome and \eqref{prop_torque} produce
\begin{align}
\bar{\tau}&=\mathop{\sum_i} \sigma_ilf_{i,p} \left[\begin{array}{c}
\cos\psi\\
\sin\psi
\end{array}\right]\nonumber \\
&-l^2D_v\left[\begin{array}{cc}
\cos^2\psi&\sin\psi\cos\psi\\
\sin\psi\cos\psi&\sin^2\psi 
\end{array}\right]\dot{\mathbf{\xi}}.\nonumber \\
&=\bar{\tau}_p+\bar{\tau}_d\label{eq:torque_gyro}
\end{align}
Since the first term in \eqref{torque_gyro} is dependent on the propelling thrusts and the second term is a linear function of $D_v\dot{\xi}$, they are referred to as $\bar{\tau}_p$ and $\bar{\tau}_d$. Equations~(\ref{eq:EoM}) and (\ref{eq:torque_gyro}) describe the reduced attitude dynamics of the bicopter near its relaxed hovering state. Together, they state that the dynamics of $\hat{\mathbf{z}}_b$ ($\mathbf{\xi}$) depends on the propellers' thrust ($f_i$'s).
\if
However, as mentioned above, when the angle between $\hat{z}_{b}$
and $\hat{z}_{w}$ is small, $\tau_{x}$ and $\tau_{y}$ in \eqref{tauR}
can be approximated by 
\begin{equation}
\mathbf{T}=r\left(f_{1}-f_{2}\right)\left[\begin{array}{c}
\cos\theta\\
\sin\theta
\end{array}\right],\label{eq:tautoF}
\end{equation}
where angle $\theta$ represents the instantaneous yaw angle which
can be approximated by the angle between $\hat{x}_{b}$ and $\hat{x}_{w}$
as illustrated by \Figref{Schematics} (B).

Term $-b_{s}\dot{\mathbf{\Xi}}$ in \eqref{EoM} is a linear-dependent\textcolor{black}{{} }stationary
damping term \cite{Crandall1985effect} which is introduced to model
the torque caused by rotor drag when the propellers are moving up
and down due to non-zero $\dot{\mathbf{\Xi}}$. $b_{s}$ is the lumped coefficient depending on $k_{\bot}$ and $r$.

The linear $\dot{\mathbf{\Xi}}$-dependent damping torque is computed
based on following derivation.

we suppose the spinning robot has infinite propellers equally distributed
at a cycle with radius $r$ around the center of mass. The vertical
rotor drag coefficient of each propeller are same and equal to $k_{\perp}^{\prime}$. 

When the robot has an attitude changing rate $\dot{\mathbf{\Xi}}$.
We define a unit 2D vector $\mathbf{V}\left(\phi\right)$ which is
rotate vector $\dot{\mathbf{\Xi}}$ an angle $\phi$ in clockwise
direction ($\phi\in\left[-\pi,\pi\right]$). Hence, we can use $\phi$
to determine propeller located at $\mathbf{V}$. So the resulted damping
torque is compute by summation of all of damping force of propellers
times the its moment arm. For $\phi$ from $-\pi$ to $0$, the propeller
are moving upward. For $\phi$ from $0$ to $\pi$, the propeller
are moving downward. So the resulted damping torque is a vector with
magnitude of 
\begin{align}
\left\Vert \mathbf{T}_{d}\right\Vert  & =2\int_{0}^{\pi}k_{\perp}^{\prime}\left\Vert \dot{\mathbf{\Xi}}\right\Vert \sin\phi d\phi\\
 & =4k_{\perp}^{\prime}\left\Vert \dot{\mathbf{\Xi}}\right\Vert \propto\left\Vert \dot{\mathbf{\Xi}}\right\Vert 
\end{align}

and direction of $-\dot{\mathbf{\Xi}}$. Therefore the damping torque
can be modeled by $-b_{s}\dot{\mathbf{\Xi}}$.
\fi

\subsection{Flight control\label{subsec:Control-Method2}}

With the description of the translational and reduced attitude dynamics, in this section, we propose to control the bicopter flight in a cascaded manner. First, \eqref{transdynamics2} is employed to determine the reference robot's attitude and total thrust that minimize the position error. Then, the attitude controller evaluates the torque required to realized the desired attitude. Lastly, a low-level controller computes the cyclic motor thrust that would generate the desired torque taking into account the gyroscopic motion of the bicopter.


\subsubsection{Position control\label{subsec:Position-Control-1}}

The position controller directly leverages the model of the translational dynamics provided by \eqref{transdynamics2} and the reduced attitude state. Given the desired trajectory $\mathbf{p}_{d}$, the control law used to compute $\mathop{\sum_{i}}f_{i,p}$ and the desired attitude state ($\mathbf{\xi}_d=[\xi_{x,d},\xi_{y,d}]$) is
\begin{align}
  \mathop{\sum_{i}}f_{i,p}\left[\begin{array}{c}
\xi_{y,d}\\
-\xi_{x,d}\\
1
\end{array}\right]&=\ddot{\mathbf{p}}_d-\mathbf{K}_{p,d}\dot{\tilde{\mathbf{p}}}-\mathbf{K}_{p,p}\tilde{\mathbf{p}}-\mathbf{K}_{p,i}\int\tilde{\mathbf{p}}\mathrm{d}t\nonumber \\
 & +mg\mathbf{e}_3\label{eq:Positioncontrol2}
\end{align}
where $\tilde{\mathbf{p}}=\mathbf{p}-\mathbf{p}_{d}$ is the position error, $\mathbf{K}_{i}$'s  are diagonal positive gain matrices. Under the assumption that the closed-loop attitude dynamics are sufficiently fast ($\hat{z}_b=[\xi_{y,d},-\xi_{x,d},1]^\mathrm{T}$), the controller guarantees the stability as $\ddot{\tilde{\mathbf{p}}}+\mathbf{K}_{p,d}\dot{\tilde{\mathbf{p}}}+\mathbf{K}_{p,p}\tilde{\mathbf{p}}+\mathbf{K}_{p,i}\int\tilde{\mathbf{p}}\mathrm{d}t=0$.

\subsubsection{Attitude control\label{subsec:Attitude-Control}}

The role of the attitude controller is to stabilize the flight (keep the robot approximately upright) and simultaneously realized the desired attitude commanded by the position controller. Owing to the complexity of the attitude dynamics, this is achieved under several simplifying assumptions. To begin, presuming that the robot is capable of generating the desired torque along the horizontal plane in the inertia frame (such that $\bar{\tau}_p=\bar{\tau}_{p,d}$) we consider a hypothetical PD controller

\begin{equation}
\bar{\tau}_{p,d} =-\delta_i K_{\tau,p}\left[\begin{array}{cc}
0 & 1\\
-1 & 0
\end{array}\right]\tilde{\xi}-K_{\tau,d}\dot{\tilde{\xi}}, \label{eq:AttitudeController}
\end{equation}
where $\tilde{\xi}=\xi-\xi_d$. With the assumption that the attitude dynamics are substantially faster than the translational dynamics, $\xi_d$ is treated as a constant or $\dot{\xi}_d,\ddot{\xi_d}=0$. Using \eqref{EoM}, the control law brings about the following closed-loop dynamics:
\begin{align}
I_{d}\ddot{\tilde{\mathbf{\xi}}}& + \left(I_{z}\Omega^*_{z}\left[\begin{array}{cc}
0 & 1\\
-1 & 0
\end{array}\right]
+K_{\tau,d}\right)\dot{\tilde{\xi}} \nonumber \\ &+\delta_iK_{\tau,p}\left[\begin{array}{cc}
0 & 1\\
-1 & 0
\end{array}\right]\tilde{\xi}=\bar{\tau}_d.\label{eq:EoM_closed_loop}
\end{align}
Since $\bar{\tau}_d$ is dependent on $\psi$, which is time-varying, the stability property of the system cannot be readily obtained by exploiting the analysis for linear time invariant (LTI) systems. Nevertheless, in cases with $|\Omega_z|=|\dot{\psi}| \gg |\dot{\xi}|$ (the yaw rate prevails) owing to the notable yaw torque in the near hovering condition, $\bar{\tau}_d$ can be approximated as its cycle-averaged value, or $\bar{\tau}_d\approx -\sfrac{1}{2}l^2D_v\dot{\xi}$ (from \eqref{torque_gyro}). \Eqref{EoM_closed_loop} becomes a two-dimensional second-order LTI system of which the stability conditions can be evaluated with the Routh-Hurwitz stability criterion. The conditions are
\begin{equation}
K_{\tau,d}+\frac{1}{2}l^{2}D_{v}>\frac{\delta_{i}I_{d}K_{\tau,p}}{I_{z}\Omega_{z}^{*}}\quad\text{and}\quad\delta_{i}K_{\tau,p}>0.\label{eq:condition}
\end{equation}
Note that according to \eqref{attitude_equilibrium}, the sign of $\Omega^*_z$ is determined by $\delta_i$. This necessitates the inclusion of $\delta_i$ in the controller to ensure that $\delta_i K_{\tau,p}/I_z\Omega_z^* > 0$. The criteria suggests that the derivative term in \eqref{AttitudeController} plays an identical (damping) role to the rotor's drag, whereas the existence of $K_{\tau,p}$ is also vital to the stability as it deals with the gyroscopic effect (the term with $\Omega_z^*$). The presence of $K_{\tau,d}$ is not absolutely needed to satisfy the first condition in \eqref{condition} as long as $K_{\tau,p}$ remains sufficiently small and the rotor drag term $l^2D_v$ is adequately significant.

\subsubsection{Torque generation\label{subsec:Time-Varying-Control-Signal}}

The stability analysis of the closed-loop attitude dynamics above has employed the assumption that the desired torque $\bar{\tau}_{p,d}$ can be realized from $\bar{\tau}_p=\mathop{\sum_i} \sigma_i lf_{i,p}[\cos\psi,\sin\psi]^\mathrm{T}$. This, however, cannot be completely achieved due to the bicopter being severely underactuated. With two independent inputs $f_i$'s, the outlined position control (\eqref{Positioncontrol2}) imposes one constraint ($\mathop{\sum_i}f_i$) on the value of $f_i$'s, leaving only one degree of freedom (DOF) for the desired 2-DOF condition $\bar{\tau}_{p}=\bar{\tau}_{p,d}$.

To workaround the underactuation, the following strategy is proposed to render the condition $\bar{\tau}_{p}=\bar{\tau}_{p,d}$ fulfilled on a cycle-average basis. This is achieved by enforcing the condition
\begin{equation}
\mathop{\sum_i} \sigma_i lf_{i,p}=2\left(\bar{\tau}_{p,d} \cdot[\cos\psi,\sin\psi]^\mathrm{T} \right ),\label{eq:LowLevelLaw}
\end{equation}so that
\begin{align}
\bar{\tau}_{p}&=2\left(\bar{\tau}_{p,d} \cdot\left[\begin{array}{c}
\cos\psi\\
\sin\psi
\end{array}\right] \right )\left[\begin{array}{c}
\cos\psi\\
\sin\psi
\end{array}\right]\nonumber \\
&\approx \bar{\tau}_{p,d} +\sin{2\psi}\left[\begin{array}{cc}
0 & 1\\
1 & 0
\end{array}\right]\bar{\tau}_{p,d}= \bar{\tau}_{p,d}+\Delta \bar{\tau}_{p,d},\label{eq:course1}
\end{align}
where we have taken $\bar{\tau}_{p,d} $ to be slowly time-varying with respect to $\psi$ (equivalent assumption to $|\dot\xi |\ll| \Omega^*_z|$). $\Delta \bar{\tau}_{p,d}$ is assigned to represent the leftover term. Since the cycle average value of $\sin{2\psi}$ is zero, the strategy described by \eqref{LowLevelLaw} ensures that the cycle-average values, denoted by $\left < \cdot \right >$, of $\Delta \bar{\tau}_{p,d}$ vanishes.  Hence, $\left < \bar{\tau}_{p}\right >=\left < \bar{\tau}_{p,d}\right >$ as intended.

\subsection{Practical considerations}

\begin{figure}
\centering{}\includegraphics{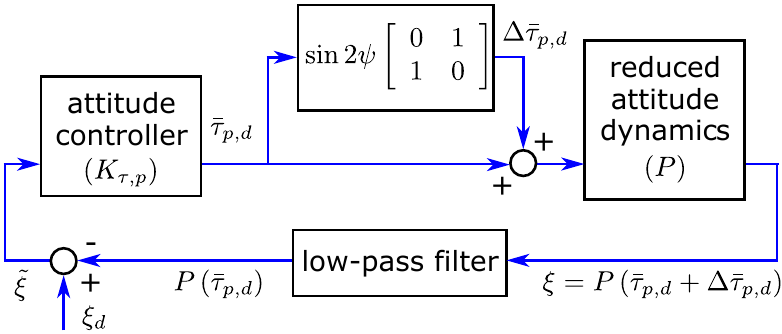}\caption{A block diagram showing the principles of the reduced attitude dynamics, attitude controller, and the low-pass filter implementation.
\label{fig:block_diag}}
\end{figure}

The proposed flight controller with the cascaded structure relies on fundamentally assumptions associated with i) the difference in timescales of the translational dynamics and the attitude dynamics; and ii) the gyroscopic motion (small angle deviation   $\left|\xi_{x}\right|,\left|\xi_{y}\right|\ll1$ and fast yaw rotation $| \Omega^*_z|\gg |\dot\xi |$). The first condition is commonly used in control of multirotor vehicles and can be easily satisfied in non-aggressive flights. To ensure other conditions are valid, following customizations are implemented

\subsubsection{Modification to the position controller}
The position control law derived from \eqref{Positioncontrol2} produces a setpoint ($\xi_d$) for the attitude controller. In practice, the magnitude of $\xi_d$ is saturated to $\xi_d^\dagger = 0.12$ rad to ensure that $\xi$ remains sufficiently small.

\subsubsection{Modification to the attitude controller\label{subsec:High-Frequency-Oscillation}}
Two primary adjustments are applied to the attitude controller to deal with the assumption related to the derivation of \eqref{course1} that $\bar{\tau}_{p,d} $ is slowly time-varying and the term $\Delta \bar{\tau}_{p,d}$  is negligible.  

First, to ensure that $\bar{\tau}_{p,d} $ does not change rapidly, we revisit the stability analysis of the PD controller given by \eqref{AttitudeController}. In experiments, we omit the $K_{\tau,d}\dot{\tilde{\xi}}$ term as it has been found inessential as demonstrated by \eqref{condition}. The elimination of $\dot{\tilde{\xi}}$ from $\bar{\tau}_{p,d}$ is beneficial as the actual torque generated by the robot ($\bar{\tau}_p$) contains a high frequency term ($\Delta \bar{\tau}_{p,d}$ in \eqref{course1}), which would induce the high frequency oscillation, rendering $\dot{\tilde{\xi}}$ to have a high frequency component in closed loop.

In addition, $\tilde{\xi}$  in  the proportional term in \eqref{AttitudeController}) is also modified as 
\begin{align}
 \tilde{\xi}\rightarrow\begin{cases}
0, & \text{if \ensuremath{|\tilde{\xi}|<\tilde{\xi}^{\dagger}}}\\
\tilde{\xi}^{\dagger}\frac{{\displaystyle \tilde{\xi}}}{{\displaystyle |\tilde{\xi}|}} & \text{otherwise},
\end{cases}
\end{align}
where $\tilde{\xi}^\dagger=0.035$ is a scalar threshold value. The respective piecewise function can be perceived as a bang-bang controller with a deadzone. Instead of a linear function, the bang-bang implementation makes sure that $\left|\bar{\tau}_{p,d}\right|$ becomes  constant except for the switching region. The inclusion of the deadzone then alleviates possible chattering effects when the error ($\tilde{\xi}$) is small. Together, they make $\left|\bar{\tau}_{p,d}\right|$ largely constant while retaining the characteristic of the proportional control.

The second adjustment is related to the existence of $\Delta \bar{\tau}_{p,d}$ in \eqref{course1}. As illustrated by \Figref{block_diag}, the term $\Delta \bar{\tau}_{p,d}$ can be regarded as input disturbance to the reduced attitude dynamics. If $P$ is employed to represent the linear plant ($\xi=P\bar{\tau}_p$, derived primarily from equation \eqref{EoM}) , the system output is $\xi=P\bar{\tau}_{p,d}+P\Delta\bar{\tau}_{p,d}$. The attitude controller subsequently takes a measurement of $\xi$ to compute the feedback $\bar{\tau}_{p,d}$ in closed loop. 

According to a linear system analysis, the disturbance term $\Delta\bar{\tau}_{p,d}$, of which the amplitude depends on $\bar{\tau}_{p,d} $, should not adversely affect the stability as long as the plant's bandwidth is smaller than the dominant frequency of $\Delta\bar{\tau}_{p,d}$ or $\Omega^*_z$  and the controller gain ($K_{\tau,p}$,  computed from \eqref{AttitudeController} given that $K_{\tau,d}=0$) is sufficiently large, making the closed-loop gain  (nominally $P/(1+K_{\tau,p}P)$) at high frequency less than unity.  However, the preference for a large $K_{\tau,p}$ contradicts the requirement of small $K_{\tau,p}$ imposed by \eqref{condition} when $K_{\tau,d}$ is set to be zero. To resolve the conflicting requirements, we incorporate a simple low-pass filter into the feedback loop as shown in \Figref{block_diag} to specifically attenuate the high-frequency component caused by $\Delta\bar{\tau}_{p,d}$ from the proposed torque generation method.

\section{Quadcopter Mode}

When two bicopters are rigidly attached as shown in \Figref{design}B, they constitute an aerial vehicle resembling a conventional quadrotor, but with multiple batteries, two flight controllers, and two pair of motors located on different horizontal planes. These differences, nonetheless, do not directly differentiate the flight dynamics of SplitFlyer in the quadcopter configuration from a regular multirotor robot as the displacement of the propellers along the vertical axis does not affect the attitude dynamics \cite{mu2019universal}. Furthermore, two bicopter modules are vertically placed 45 mm apart, almost twice the propellers' radius (27.5 mm), to reduce the possible airflow interruption and deterioration in aerodynamic efficiency.

Since the topic of flight dynamics and control of a conventional multirotor vehicle is beyond the scope and not a contribution of this work, we incorporate a standard cascaded controller \cite{mellinger2012trajectory} for the  attitude and position control loops. \textcolor{black}{An identical flight controller for the quadcopter mode is deployed on both bicopter robots. Without direct communication, each bicopter commands its two propellers according to the quadcopter control law based on its own IMU measurements while assuming the other robot half behaves in a similar manner. Therefore, the behavior of the entire quadcopter is theoretically identical to that of a regular quadcopter will a single control board.} With no consideration of rotor drag, this implementation benefits from its simplicity and low computation, at the sacrifice of tracking performance when it comes to more aggressive maneuvers.

\section{Experimental Validation}
\subsection{SplitFlyer prototypes}
For construction of the bicopter modules, we chose the all-in-one flight control board from Crazyflie 2.0 (Bitcraze) thanks to the ability to easily modify the low-level controller. The structural components were fabricated from 3D printed parts (Formlabs Form 2) and carbon fiber rods in an attempt to minimize the weight. For propulsion, $7\times20$-mm coreless DC motors and propellers with 27.5-mm radius were selected. Two propellers were placed 120 mm apart.

To ensure that the bicopter approximately behaves as a gyroscope as assumed during the modeling of flight dynamics, two 100mAh 1s Li-ion batteries were installed on each bicopter, diagonal from the propellers as seen in \Figref{design}A. \textcolor{black}{The battery and motor weigh 2.8 g and 2.9 g each. They are placed 90 mm and 59 mm away from the center.} This provides the desired mass distribution such that the moment of inertia about the roll ($\hat{x}_b$) and pitch ($\hat{y}_b$) axes of the robot are $6.0\times 10^4 $ and $5.5\times 10^4$ g$\cdot$mm$^2$  or within 10$\%$  of each other as estimated by CAD software (Fusion 360), while the inertia about the yaw axis is $11.3\times 10^4$ g$\cdot$mm$^2$. \textcolor{black}{Possible damages on the exposed batteries from a collision is unlikely thanks to the vehicle's small mass.} The total mass of the bicopter is 26.0 g. When combined, the SplitFlyer weighs 55.1 g, including 3.1 g from the attachment mechanism.

\subsection{Experimental setup and flight arena}
Flight experiments were carried out in a $3\times3\times2.5$-m
arena equipped with six motion capture cameras (OptiTrack Prime 13w) for tracking the position and orientation of the vehicles for flight
control and ground-truth measurements. 

Both Bicopter-CW and Bicopter-CCW were programmed with flight controllers for both flight modes, leveraging the officially provided source code. The controller is prescribed to monitor the vehicle's yaw rate and automatically executes a suitable flight mode, switching to a bicopter flight controller when detecting $|\dot{\psi}|>8.7$ rad$\cdot$s$^{-1}$ from the built-in gyroscope.

Communication between the robots and the ground station was achieved with Bitcrazy Crazyradio PA. In the quadrotor mode, the SplitFlyer received the position and yaw feedback from the motion capture via the ground station. Both control boards on the robot functioned independently, controlling one pair of motors each. The IMU feedback provided by both boards were nearly identical and proved not to weaken the vehicle's stability. 

In the bicopter form, both position and attitude controllers were implemented on the ground station and executed at 100 Hz. This directly used the position and $\xi$  feedback from the motion capture system. The desired torque ($\bar{\tau}_{p,d}$ ) was transmitted to the robot with the instantaneous yaw angle ($\psi$). The flight control board generated the motor commands from $\bar{\tau}_{p,d}$ and $\psi$ at 200 Hz. The yaw angle $\psi$  used was  obtained by fusing the integrated gyroscope measurement with the motion capture feedback to ensure a fast update rate with continuous correction.

\subsection{Flight experiments}

\subsubsection{Quadcopter Flight}
\begin{figure}
\begin{centering}
\includegraphics{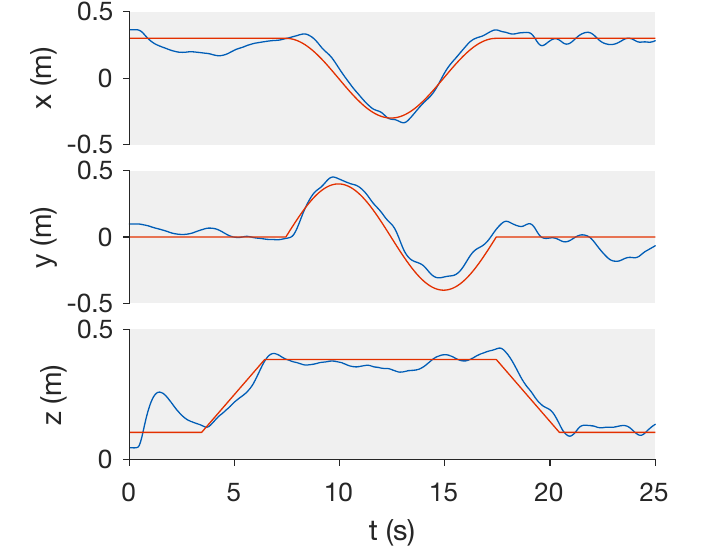}
\par\end{centering}
\caption{Trajectory of the robot in the quadrotor mode. Blue lines are measurements and red lines are references.
\label{fig:quadexpdata}}
\end{figure}
\begin{figure}
\begin{centering}
\includegraphics{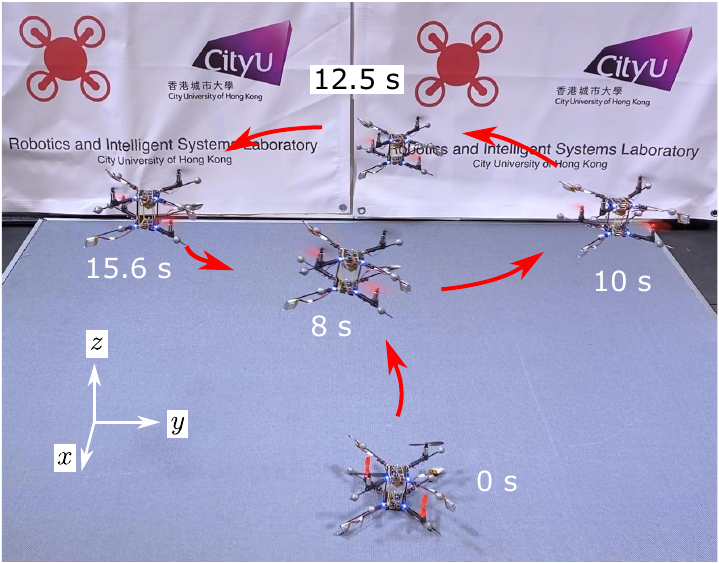}
\par\end{centering}
\caption{A composite image taken from the flight experiment of the robot flying in the quadrotor configuration. \label{fig:quadexp}}
\end{figure}

For demonstration, the SplitFlyer was reconfigured as a quadcopter. As previously stated, two flight control modules received the same position and yaw feedback from the ground station. The robot's attitude was controlled by both controller boards with no communication in-between. We constructed a simple 25-s trajectory consisting of takeoff, hovering, elliptical and landing phases.  

Using a standard cascaded flight controller \cite{mellinger2012trajectory}, we conducted one flight test to verify the flight capability in this mode. The resultant trajectory is shown in \Figref{quadexpdata} alongside the reference. A composite image constructed from a flight video is presented in \Figref{quadexp}.  They verify that the quadcopter constructed from two bicopter modules produced a satisfactory flight performance, with the root-mean-square (RMS) errors in horizontal and vertical directions of $9.7$ and $7.6$ cm. The magnitudes are reasonable given the simple implementation of the controller.
\subsubsection{Bicopter Flight}
\begin{figure}
\begin{centering}
\includegraphics{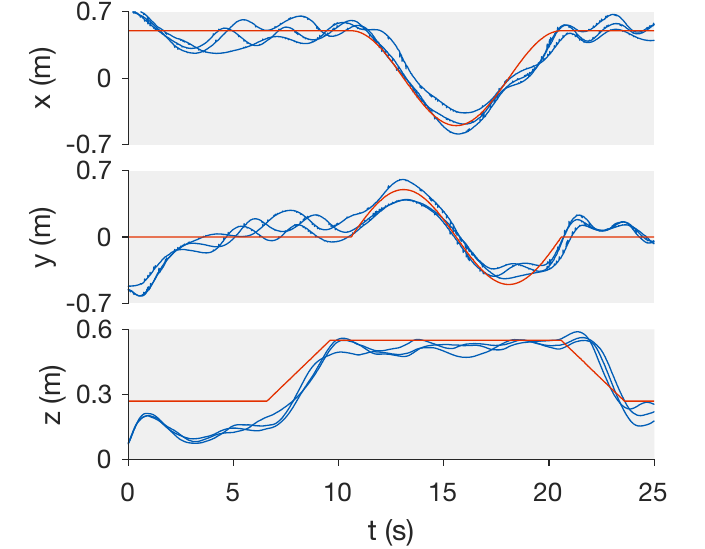}
\par\end{centering}
\caption{Trajectories of the robot in the bicopter mode. Three flights are presented. Blue lines are measurements and red lines are references. \label{fig:quadexpBidata}}
\end{figure}
\begin{figure}
\begin{centering}
\includegraphics{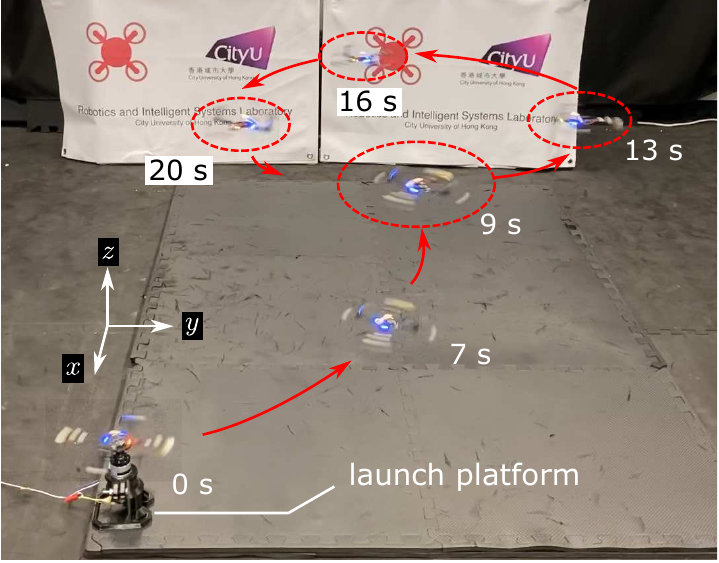}
\par\end{centering}
\caption{A composite image taken from the flight experiment of the robot flying in the bicopter configuration. \label{fig:quadexpBi}}
\end{figure}
\begin{figure}
\begin{centering}
\includegraphics{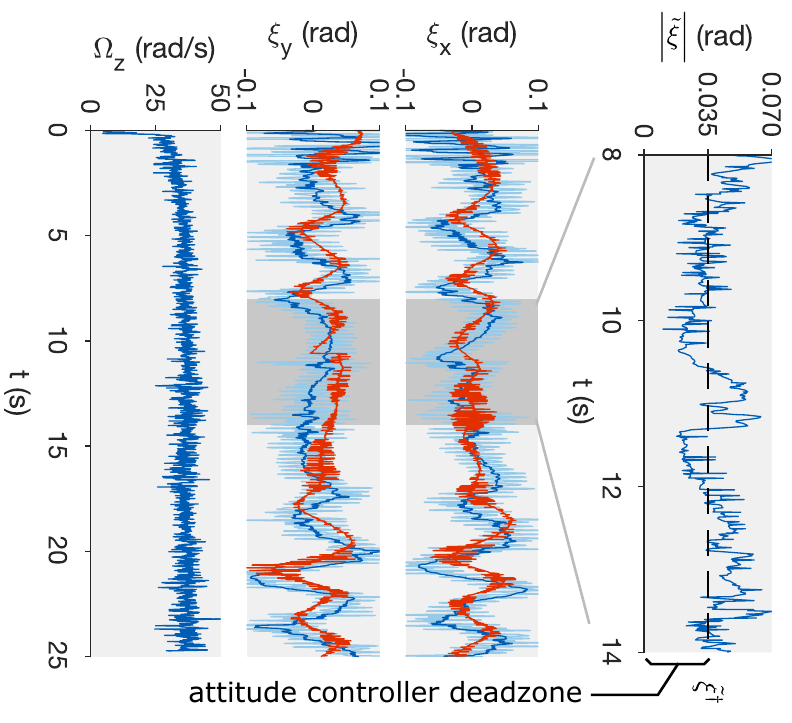}
\par\end{centering}
\caption{Plots of experimental data taken from one of the bicopter flight. The top plot is a close-up of the attitude state error with the controller deadzone. The second and third plots are the attitude state in terms of $\xi_{x}$ and $\xi_{y}$. Light blue lines are raw measurements, dark blue lines are filtered data and red lines are the references ($\xi_{x,d},\xi_{y,d}$ ) provided by the position controller. The bottom plot shows the yaw rate of the robot during flight.
 \label{fig:quadexpBidata2}}
\end{figure}
For a bicopter, a similar trajectory was chosen to verify the flight stability and trajectory following capability. The elliptical phase was replaced with a circular path. To allow the robot to initialize with the bicopter flight mode, a motorized rotating launch platform was used to provide the robot the  initial yaw speed \textcolor{black}{(approximately 10 rad$\cdot$s$^{-1}$, higher the controller switching yaw rate of 8.7 rad$\cdot$s$^{-1}$)}. Three 25-s flights were carried out. The trajectories recorded by the motion capture feedback are provided in \Figref{quadexpBidata}. A composite photo taken from an example flight is shown in \Figref{quadexpBi}.

\Figref{quadexpBidata} confirms that the proposed strategy is capable of stabilizing the robot and providing acceptable trajectory following performance. In all three flights, the robot evidently tracked the reference trajectory with the RMS errors of $15.7$ cm and $8.8$ cm in horizontal and vertical directions. The errors are slightly larger than those from the quadcopter mode. The likely explanation is the degree of underactuation of the vehicle, which allows the robot to realize the desired torque only on a cycle-average basis.

\Figref{quadexpBidata2} provides further flight data. The x and y components of $\hat{z}_b$ provided by the motion capture system are plotted as $\xi_x$ and $\xi_y$ with their respective setpoints ($\xi_{x,d},\xi_{y,d}$) from the position controller. It can be seen the primary frequency of $\xi_d$ is $\approx$ 1/3 Hz (notably slower than $\Omega_z$, which is $\approx35$ rad$\cdot$s$^{-1}$). This reflects the timescale of the translational dynamics (consistent with the data in \Figref{quadexpBidata}). The filtered measurements of $\xi$ are $\approx 1$ s behind the setpoints, suggesting that the closed-loop attitude dynamics are not substantially faster than the translational dynamics. This prevents the robot from tracking the desired position more accurately. 

The top of \Figref{quadexpBidata2}  shows a close-up view of $\tilde{\xi}$ from one of the flights. This represents the feedback of the attitude controller. As discussed in Section \ref{subsec:High-Frequency-Oscillation}, the control torque is only generated when $|\tilde{\xi}|>\tilde{\xi}^\dagger=0.035$. Over three flights, it was found that the control torque was enabled in $68\%$ of the flight period. This further explains the limitation  of the proposed strategy that balances the flight stability against the tracking performance.

\subsection{Sequential quadcopter and bicopter flights}
To highlight the transformation and flight mode switching, we carried out a continuous experiment involving both robot configurations. Starting in the quadcopter mode, the robot flew briefly and landed. A human operator then manualy took the robot apart. Two bicopters were placed on the launch platforms and commanded to takeoff consecutively. Both bicopters simultaneously demonstrated stable flights in the arena for over 20 s and safely landed afterwards. The video of the total process is provided as a supplement to this paper.

Furthermore, we qualitatively verified that a launch platform is, in fact, not mandatory for initializing the bicopter flight. The video attachment shows the robot robustly stabilizes its attitude and position after being hand-thrown by an operator. \textcolor{black}{ the throwing motion induced the initial angular velocity of 26 rad$\cdot$s$^{-1}$, close to the equilibrium revolving speed of the bicoper (see \Figref{quadexpBidata2})}. The robot then recovered and stabilized from the significant initial velocity.

\section{Conclusion and Future Work}
In this work, we have developed a transformable aerial vehicle--SplitFlyer. In addition to a regular quadcopter flight mode, SplitFlyer seprates into two flight-capable bicopters that can function independently. The two modes of aerial locomotion are vastly different. In the bicopter configuration, the robot, possessing only two actuators, is severely underactuated. Yet, it has been shown controllable in terms of its position and attitude with some relaxation on the yaw state. The proposed cascaded control strategy was experimentally verified. Finally, we demonstrated a consecutive conceptual flight test. Starting from a quadcopter flight, the SplitFlyer was reconfigured into bicopters with human assistance. Both bicopters then simulaneously lifted off and flew independently.

This work could be considered an important milestone towards a modular aerial robot that can autonomously disassemble into multiple agents mid-flight. Future work will focus on an automated detachment mechanism, with a possibility of mid-air re-docking.

\bibliographystyle{IEEEtran}
\bibliography{BibTex/bibtex}

\end{document}